\pdfoutput=1

\documentclass[11pt]{article}

\usepackage{acl}

\usepackage{times}
\usepackage{latexsym}
\usepackage[T1]{fontenc}
\usepackage[utf8]{inputenc}
\usepackage{microtype}
\usepackage{inconsolata}
\usepackage{graphicx}
\usepackage{caption}
\usepackage{amsmath}
\usepackage{amssymb}
\usepackage{comment}
\usepackage{booktabs}
\usepackage{hyperref}
\usepackage[overload]{empheq}
\usepackage{makecell}
\usepackage{xcolor,colortbl}
\usepackage[ruled,vlined]{algorithm2e}
\usepackage{array,multirow}
\usepackage{comment}
\usepackage{pifont}
\usepackage{siunitx}
\usepackage{url}
\usepackage{enumitem}
\usepackage{arydshln}
\usepackage{adjustbox}
\usepackage{xcolor}
\usepackage{soul}

\usepackage{tabularx}
\usepackage{ragged2e}
\usepackage{arydshln}
\usepackage{changepage}
\usepackage{listings}
\usepackage{diagbox}

\newcolumntype{Y}{>{\RaggedRight\arraybackslash}X}

\title{When Words Divide: Diachronic Ideological Polarization in \\Political Discourse on Social Media}

\author{
Roy Yitzchak \hspace{2cm}
Noa Lavie \hspace{2cm}
Ella Rabinovich \\
The Academic College of Tel-Aviv Yaffo, Israel \\
\texttt{\{royyi,lavie,ellara\}@mta.ac.il} \\
}

\begin{document}
\maketitle
\begin{abstract}
Political polarization has become a defining feature of online discourse, yet its long-term evolution remains poorly understood. We present a longitudinal analysis of ideological polarization in Reddit discussions by measuring semantic differences in the language used by opposing political communities. We construct temporally aligned community-specific word embeddings and quantify ideological polarization as the semantic divergence of political concepts over time. Our analysis shows that ideological polarization has increased substantially during the study period, both at the concept- and topic-level.
Unlike prior computational work, which has largely focused on cross-sectional analyses or affective dimensions of polarization at a single point at time, our approach captures the evolution of ideological differences in semantic framing. The proposed framework provides a scalable method for studying the temporal dynamics of ideological polarization in large-scale social media discourse.

\end{abstract}

\section{Introduction}
\label{sec:introduction}

Political polarization has become an increasingly prominent phenomenon over the past couple of decades; among the proposed reasons are social media, as well as other societal and technological changes. Both theoretical and computational approaches to political science have studied polarization along two complementary dimensions: (1) \textit{ideological} polarization, which can be captured through divergences in the ways topics are discussed, and (2) \textit{affective} polarization, which reflects the emotions conveyed through language by parties with opposing political opinions.

Multiple studies have examined differences between political wings' points of view through the lens of language; most works present a comparative analysis at a single point in time, in news outlets \citep{sinno2022political}, U.S. congressional speeches \citep{jensen2012political}, and on social media \citep{conover2011political, demszky2019analyzing, milbauer2021measuring}. However, both settings have drawbacks: the former typically introduces a "clean", post-edited form of language suited for the press, while the latter, often based on data collected when Twitter posts were limited in length, focuses on an extreme example of social media dialect, characterized by slang, abbreviations, emojis, and other platform-specific conventions.
Additionally, very little work has been devoted to longitudinal diachronic analyses of polarization, with the notable exception of \citet{jensen2012political}, who examined the evolution of political polarization in the United States by analyzing partisan language in congressional speeches over more than 130 years, and the more recent study by \citet{goldin2026affective}, who examined diachronic \textit{affective} polarization in Israeli parliamentary proceedings over 30-year period.

Our work goes above and beyond the current state of the art in three important ways: (1)~We collect and release a longitudinal, large-scale, diverse, and high-quality dataset of authentic posts authored by thousands of users with left- or right-wing political affiliations on Reddit\footnote{\url{https://www.reddit.com/}} -- one of the most popular discussion platforms worldwide;\footnote{We use "left-wing" and "Democratic", and "right-wing" and "Republican", interchangeably for convenience, recognizing that these terms are not perfectly synonymous but broadly align in the contemporary U.S. political context.} (2)~We propose a methodology for concept-level diachronic analysis of ideological polarization and conduct analyses of textual data spanning more than a decade, and (3)~We abstract away from individual concepts and perform a higher-level analysis of topical trends, showing that increasing ideological divergence is concentrated in substantive policy and identity-related topics.

Our contributions in this study are therefore as follows: First, we collect and release a dataset containing almost 5M posts and comments (about 85M words) gathered from Reddit topical threads (subreddits) associated with Republicans and Democrats. Second, we apply both previously proposed and novel methods for studying ideological polarization in social media -- a framework that can be reused in future studies. And finally, we show reliably detected increasing polarization trends -- a finding that provides strong empirical support for theoretically motivated hypotheses on rising polarization in society. All data and code will be made available.


\section{Related Work}
\label{sec:related-work}

Political polarization has long been a central topic in political science, where it is commonly viewed as a multidimensional phenomenon encompassing both \textit{ideological} and \textit{affective} components \citep{mccarty2006polarized, iyengar2012affect}. Ideological polarization refers to increasing divergence in policy preferences, political beliefs, and issue positions between competing political groups, and has traditionally been studied using legislative voting records and related measures of political ideology \cite{poole1985spatial, mccarty2006polarized}. More recently, scholars have also studied affective polarization, which captures the extent to which partisans increasingly dislike, distrust, and express hostility toward members of opposing political groups, independently of policy disagreements \citep{iyengar2012affect, mason2018uncivil, iyengar2019origins}. Focusing on \textit{ideological} polarization in political discourse on social media, we leave the affective dimension to a follow up study.

\paragraph{Computational Approaches} Building on these theoretical foundations, computational social science and natural language processing have increasingly sought to measure political polarization directly from large-scale textual data. Early work analyzed legislative speech, showing that linguistic differences between political parties mirror ideological divisions observed in voting behavior \citep{jensen2012political, gentzkow2019measuring}. The growing availability of social media data subsequently enabled polarization to be studied in online political discourse, where researchers investigated partisan communities, ideological framing, and echo chambers at unprecedented scale \citep{conover2011political, demszky2019analyzing}. More recent studies have employed distributional semantics and contextual representations to quantify ideological differences through semantic divergence, revealing that identical political concepts may acquire different meanings across communities \citep{milbauer2021measuring, sinno2022political}. Despite substantial methodological advances, most computational studies examine polarization at a single point in time (except \citet{jensen2012political}), providing snapshots of ideological differences between communities, rather than their temporal evolution.

\paragraph{Diachronic Studies} Compared with cross-sectional analyses, relatively little computational work has examined the temporal evolution of political discourse. \citet{jensen2012political} analyzed more than a century of U.S. congressional speeches, demonstrating increasing linguistic polarization over time. Recently, \citet{goldin2026affective} proposed a computational framework for measuring \textit{affective} polarization, showing that discourse in Israel Parliament (the Knesset) has become increasingly emotionally polarized over the past three decades. The study most closely related to ours is \citet{rivlin-angert2025enemy}, who tracked \textit{delegitimization} in Israeli political discourse over time across parliamentary speeches, social media, and news by detecting language that questions the legitimacy of political opponents. While the conceptually ask "Is the speaker denying the opponent's legitimacy as a political actor?", we study how individual concepts' (e.g., "election", "radicals", "welfare") semantics diverges over time between the two American political communities: Republicans (right-) and Democrats (left-wing). 

Our work is, to the best of our knowledge, is the first to provide longitudinal analysis of semantic divergence in authentic social media discourse over more than fifteen years. We release a carefully collected, large-scale dataset, used in this study, to facilitate further research in this field.

\section{Dataset}
\label{sec:dataset}

We describe the data collection, preprocessing procedures, and provide the final dataset statistics.

\subsection{Data Collection}
All textual data used in this study were collected from Reddit, a large online discussion platform organized into thousands of user-created communities known as subreddits. Each subreddit focuses on a specific topic, interest, or ideological orientation, and discussions typically take place through posts and comments authored by users.

For the purpose of analyzing ideological polarization in political language, the dataset was constructed from subreddits associated with major political affiliations in the United States. Left-leaning communities included two subreddits: \texttt{r/democrats} and \texttt{r/liberal}, while right-leaning communities included \texttt{r/conservative} and \texttt{r/republican}.\footnote{Admittedly, political discourse on social media may be skewed toward more engaged or ideologically extreme users, potentially limiting coverage of the full political spectrum.} This design assumes that most users participating in these communities are broadly aligned with their respective political orientations. While no perfect separation can be expected, as users with opposing views may also contribute, active participation in ideologically oriented communities generally reflects a degree of affiliation with those communities. Similar assumptions have been made in prior studies involving Reddit data \citep{rabinovich2018native, goldin2018native, shem2025interplay}.

Our raw corpus consists of posts and comments collected over a period of approximately fifteen years (2008--2023). The data were collected from Reddit using the Pushshift API,\footnote{\url{https://pushshift.io}} and stored in \texttt{json} format. Data collection ended in 2023, when Reddit discontinued support for large-scale programmatic data retrieval.

\subsection{Data Preprocessing}
All collected data were partitioned by year of submission, and submissions shorter than three words were filtered out. As expected, the earlier years contain relatively little text, with progressively larger volumes in later periods.
Table~\ref{tab:data-stats} reports the distribution of the number of submissions and words per community (left, right) across the years, along with the corresponding totals. Evidently, subreddits associated with the political right contain nearly ten times more data than those associated with the political left. One plausible explanation is that between 2008 and 2022 Democratic users largely populated the general, left-leaning \texttt{r/politics} forum, leaving \texttt{r/democrats} underutilized. Conversely, conservative users migrated to, and consolidated their activity within, dedicated forums such as \texttt{r/republicans}, resulting in a substantial difference in data volume due to being outnumbered in mainstream spaces \citep{soliman2019characterization}. We note, however, that the analyses presented in the remainder of the paper remain robust despite this substantial class imbalance.

Below are two example comments posted on \texttt{r/democrats} and \texttt{r/republicans}, respectively, during 2019, taken verbatim from our data:

\begin{adjustwidth}{-0.1in}{-0.1in}
\begin{quote}
"Biden can tear people apart in debates, it's going to be fun to watch him slay the GOP time and again until he's our next president."
\end{quote}
\end{adjustwidth}

\begin{adjustwidth}{-0.1in}{-0.1in}
\begin{quote}
"There was more people dying from falling out of bed then from rifles."
\end{quote}
\end{adjustwidth}

\begin{table}[ht]
\centering
\resizebox{1.0\columnwidth}{!}{
\begin{tabular}{l|rr|rr}
\multicolumn{1}{c}{} & \multicolumn{2}{c|}{Republicans} & \multicolumn{2}{c}{Democrats} \\
year & texts & words & texts & words \\ \hline
2008 & 26 & 1K & -- & -- \\
2009 & 319 & 6K & 8 & 156 \\
2010 & 1,607 & 33K & 53 & 1K \\
2011 & 12,285 & 257K & 1,899 & 37K \\
2012 & 69,171 & 1,393K & 7,112 & 137K \\
2013 & 52,227 & 1,012K & 10,777 & 221K \\
2014 & 51,769 & 1,019K & 11,525 & 240K \\
2015 & 69,068 & 1,309K & 13,972 & 279K \\
2016 & 168,060 & 3,181K & 31,041 & 600K \\
2017 & 190,831 & 3,777K & 23,486 & 452K \\
2018 & 192,625 & 3,550K & 42,587 & 784K \\
2019 & 296,014 & 5,287K & 41,317 & 754K \\
2020 & 1,189,921 & 21,314K & 68,230 & 1,182K \\
2021 & 1,082,215 & 19,217K & 67,796 & 1,191K \\
2022 & 847,456 & 14,419K & 78,069 & 1,334K \\
2023 & 155,508 & 2,698K & 21,002 & 350K \\
\hline
total & 4,379,102 & 78,473K & 418,874 & 7,562K \\

\end{tabular}
}
\caption{Submission (post or comment) and word count distribution over the years in our data. Both Republican and Democratic subreddits were established in 2008.}
\label{tab:data-stats}
\end{table}

\section{Ideological Polarization}
\label{sec:ideological}

Our methodology consisted of several sequential steps: (1) identifying a set of words associated with political discourse; (2) partitioning the entire time span into consecutive periods, each containing a substantial amount of data; (3) learning \textit{contextual representations} of the politically associated words for both right- and left-leaning communities at each time point; and (4) computing the \textit{semantic divergence} of each word between the two communities at each time point, as well as its divergence trend over time. Concepts exhibiting a systematic increase in semantic divergence over time constitute the primary focus of our study.
Next, we provide a detailed description of each step.

\subsection{Identification of Political Vocabulary}
\label{ssec:political-vocab}
We identified a political lexicon characteristic of Reddit discussions by applying the log-odds ratio with informative Dirichlet prior method \citep{monroe2008fightin}, comparing an equally sized subset of left- and right-leaning communities against a neutral background corpus. We then manually inspected 1,017 words that exceeded a strict log-odds threshold of 2.0 and appeared at least 100 times in the dataset. This process resulted in a final set of 850 words with a clear political association.

Among the highest-scoring words (i.e., those exhibiting the strongest political association) were terms directly related to political identity and activity, such as \textit{democrats, conservative, president, election, republicans, liberal, campaign}. The resulting lexicon also included words that frequently occur in political discourse despite lacking an explicit political orientation, such as \textit{economy, violence, illegal, healthcare, immigrants, climate,} and \textit{education}. The full list of 850 terms is released as part of the data accompanying this study.

\subsection{Construction of Time Periods}
In order to examine semantic divergence over time, the dataset was partitioned into 11 time periods (rather than the original 15) spanning 2008--2023. This configuration was chosen to balance temporal resolution with the need for sufficiently large corpora in each period to enable reliable training of contextual word representations. In particular, data from the earlier years (2008--2011) were aggregated to ensure adequate coverage, as small corpora do not yield stable semantic representations.

\subsection{Learning Word Meaning Representation}
Semantic word representations (embeddings) learned by training word2vec \citep{mikolov2013efficient} were adopted as the primary methodological tool in this study. Semantic divergence between the representations of the 850 politically associated concepts was then computed across the two communities over the 11 time periods, facilitating a longitudinal analysis of ideological polarization.

\paragraph{Word2vec Hyperparameters}
The semantic representation training procedure is sensitive to several hyperparameters, including embedding dimensionality and context window size. To ensure that the embeddings were trained under appropriate settings, we tuned these values using the Simlex-999 benchmark dataset \citep{hill2015simlex} and explicitly validated their ability to capture genuine semantic similarity. Simlex-999 is a human-annotated dataset designed specifically to measure true semantic similarity between word pairs rather than simple association. Each word pair in the dataset was assigned a similarity score by human annotators, reflecting the degree of semantic similarity between the two words. For example, the pair "friend--buddy" received a high similarity of 8.78 (out of 10), whereas the pair "lawyer--banker" received the substantially lower score of 1.88.

Words from Simlex-999 that occurred at least 100 times in our full corpus were considered during parameter tuning. For each candidate configuration, the semantic similarity computed by word2vec for a given word pair was compared against the corresponding similarity score in Simlex-999. Among the tested configurations, the best performance yielded a Pearson correlation of approximately 0.42 using \texttt{window\_size=2} and \texttt{vector\_size=300}, while keeping all other parameters at their default values. Considering the moderate size and domain-specific nature of the Reddit corpus, this correlation was deemed to be satisfactory and indicative of the model's ability to capture semantic relationships.

\paragraph{Learning Diachronic Embeddings}
Using the best-performing word2vec configuration, we trained word representations separately for each of the 11 time periods, focusing on the 850 politically associated words identified in Section~\ref{ssec:political-vocab}. To avoid the need for post-hoc alignment of independently trained embedding spaces, which are otherwise not directly comparable, we adopted a simple lexical transformation. Specifically, each occurrence of a target \texttt{word} was marked as either \texttt{word\_R} or \texttt{word\_D}, depending on whether it originated from right- or left-leaning data, respectively. For example, the word \texttt{patriot} was transformed into \texttt{patriot\_R} and \texttt{patriot\_D} within each temporal split. Embeddings were then trained on the combined corpus of each period, and representations of all 850 words -- separately encoded as \texttt{R} and \texttt{D} variants -- were extracted and subsequently used to compute semantic divergence over time.

\subsection{Computing Diachronic Word Meaning Divergence Patterns}
\label{ssec:methodology-diachronic}
We denote the contextual representation (embedding) of a word $w$, originating from Republican- and Democratic-leaning data at time point $t$, as $e_t^r$ and $e_t^d$, respectively. Semantic divergence between the two representations of $w$ at time $t$ is defined as the cosine distance between their embeddings:

\begin{equation} 
\texttt{SDiv}_{t}(w) = 1 - \text{cosine}(e_t^r, e_t^d) 
\label{eq:divergence} 
\end{equation}

Given a word $w$ and a sequence of time points $[1, 2, \ldots, k]$, the corresponding series of semantic divergence (\texttt{SDiv}) values is represented as $[\texttt{SDiv}_{1}(w), \texttt{SDiv}_{2}(w), \ldots, \texttt{SDiv}_{k}(w)]$. A systematically increasing pattern indicates that the meanings or contextual usages of the word are \textit{diverging} between the two communities over time, whereas a decreasing pattern suggests increasing similarity. Conversely, the absence of a clear trend indicates relative stability in the contextual usage of $w$ across the political communities over time.

We used the Mann--Kendall statistical test \citep{mann1945nonparametric} to assess the significance of divergence trends for individual words, following previous work on trend detection in diachronic analyses \citep{goldin2026affective}. 
It evaluates the null hypothesis that a sequence exhibits no monotonic trend against the alternative hypothesis that it displays either an increasing or decreasing trend. The test does not assume normality and is robust to noise, making it well suited for the analysis of semantic shift trajectories. While the test is generally considered reliable when applied to sequences containing at least eight observations,\footnote{\url{https://vsp.pnnl.gov/help/vsample/design_trend_mann_kendall.htm}} our 11-period partition satisfies this requirement. Notably, the relatively small number of time points may reduce statistical power, causing some genuine trends to remain undetected; however, it is unlikely to introduce spurious significant trends in the opposite direction.

\subsection{Cross-Community Semantic Divergence}
We first demonstrate that the proposed approach reliably detects meaningful cross-community concept-level differences by considering the entire body of data (with no temporal split), divided into right- and left-wing communities.

\begin{figure}[h!] 
\centering 
\includegraphics[width=\columnwidth]{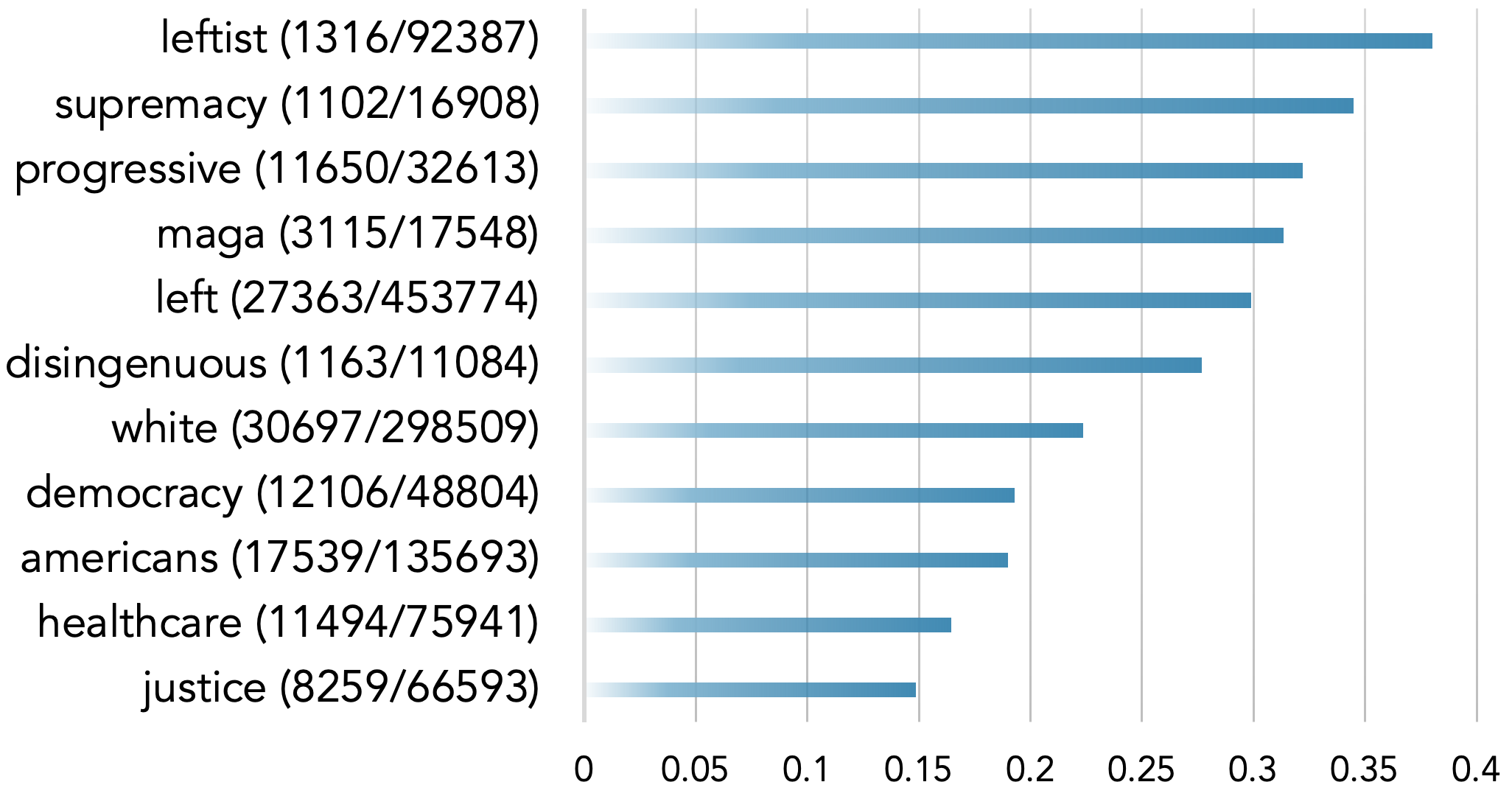} 
\caption{Example words (count in left-wing / count in right-wing) with the highest \texttt{SDiv} between the two communities: these words are systematically used in \textit{differing contexts} by Republicans and Democrats.}
\label{fig:cross-community} 
\end{figure}

\paragraph{Quantitative Analysis}
Figure~\ref{fig:cross-community} presents example words with varying degrees of semantic divergence (\texttt{SDiv}) between the two communities, considering the full 2008--2023 dataset split by political affiliation. The results show that the two political communities systematically use the same concepts in different semantic contexts.

\paragraph{Qualitative Analysis}
To further support our quantitative findings, we conducted a qualitative analysis by examining the top-K nearest neighbors of selected words in the semantic space. For each target word, we extracted its closest neighbors separately from the right- and left-leaning embedding neighborhoods, allowing us to inspect differences in contextual usage across the two communities. Table~\ref{tbl:cross-community-neigh} presents three example concepts exhibiting high divergence and one concept exhibiting low divergence between the communities.

\begin{table*}[h!]
\centering
\small

\begin{tabularx}{\textwidth}{l|c|X|X}
\multicolumn{2}{c}{} &
\multicolumn{1}{c|}{Republicans} &
\multicolumn{1}{c}{Democrats} \\

word & \texttt{SDiv} & \multicolumn{1}{c|}{top-K semantic neighbors} & \multicolumn{1}{c}{top-K semantic neighbors} \\ \hline

radicals & 0.424 &
extremists, crazies, lunatics, reformists, \textcolor{red}{marxists}, \textcolor{red}{anarchists}, whackos, \textcolor{red}{progressives}, radical, wackjobs, \textcolor{red}{leftists}, \textcolor{red}{alinskys}, \textcolor{red}{idealogues}, extremist, wackos, communists, psychos & 
xtremists, nutjobs, factions, \textcolor{blue}{reactionaries}, kooks, \textcolor{blue}{agitators}, \textcolor{blue}{antifascists}, \textcolor{blue}{islamist}, \textcolor{blue}{antigovernment}, facists, whackjobs, nutters, \textcolor{blue}{populists}, anarchists, winged, radicalism, neoliberals
\\
\hdashline

lockdown & 0.367 &
lockdowns, shutdown, shutdowns,	quarantine,	quarantines, \textcolor{red}{mandates}, shelterinplace, quarentine, \textcolor{red}{curfews}, \textcolor{red}{draconian}, selfisolation, masking, \textcolor{red}{restrictions}, stayathome, mask &
lockdowns, \textcolor{blue}{quarantine}, quarantines, \textcolor{blue}{distancing}, smallpox, \textcolor{blue}{mutations}, shutdowns, outdoors, \textcolor{blue}{precautions}, \textcolor{blue}{omicron}, airborne, hospitalizations, mask, collision, symptomatic \\
\hdashline

reparations & 0.353 &
reparation, reperations, repetitions, \textcolor{red}{descendants}, recompense, decendants, \textcolor{red}{ancestors}, \textcolor{red}{forgiveness}, \textcolor{red}{restitution}, descendents, \textcolor{red}{exslaves}, \textcolor{red}{redistribution}, refund, \textcolor{red}{slaveowners} &
\textcolor{blue}{colonialism}, chattel, freeing, \textcolor{blue}{abolition}, \textcolor{blue}{segregationists}, abolitionists, reformation, hostilities, undesirables, \textcolor{blue}{assimilation}, \textcolor{blue}{criminalization}, \textcolor{blue}{inequity}, abolishment, protectionism \\
\hline

wars & 0.143 &
war, conflicts, \textcolor{red}{quagmires}, \textcolor{red}{nationbuilding}, \textcolor{red}{adventurism}, iraq, afghanistan, ww, punic, entanglements, entangling, conflict, wwii, lybia &
war, \textcolor{blue}{unwinnable}, genocides, destabilized, revolutions, nam, iraq, \textcolor{blue}{neverending}, libya, wwi, hostilities, recessions, pullout, afganistan \\

\end{tabularx}

\caption{Examples of how the two communities attach different semantic connotations to the same concept. In the right-leaning community, "radicals" is primarily associated with ideological groups on the political left, whereas in the left-leaning community it is associated with a broader range of extremist or anti-establishment actors. During the COVID-19 pandemic, "lockdown" is framed primarily through government restrictions in the right-leaning community and through public health in the left-leaning community. The word "reparations" is embedded in the historical context of slavery and racial injustice in the left-leaning community, whereas in the right-leaning community it is associated more closely with compensation and responsibility. In contrast, "wars" exhibits relatively low semantic divergence, with both communities associating it primarily with foreign military conflicts.}

\label{tbl:cross-community-neigh}
\end{table*}

\subsection{Diachronic Semantic Divergence}
We now analyze \textit{diachronic} patterns in semantic divergence between the two communities.

\paragraph{Quantitative Analysis}
Among the 850 politically associated words, the Mann--Kendall trend test (see Section~\ref{ssec:methodology-diachronic}) identified 119 words ($\sim$14.0\%) as exhibiting \textit{increasing} semantic divergence,\footnote{The significance threshold of 0.05 was used in all tests.} 566 words ($\sim$67.0\%) as showing no significant trend over time, and only 5 words ($\sim$0.6\%) as exhibiting \textit{decreasing} semantic divergence. An additional 166 words ($\sim$18.4\%) did not occur sufficiently frequently across all 11 time periods and were therefore excluded from the analysis.
As expected, not all examined political terms exhibit statistically significant diachronic trends, as uniform semantic change across all terms is unlikely. Nevertheless, a substantial proportion of words (14.0\%) exhibit increasing semantic divergence over time, compared to only 0.6\% exhibiting a decreasing trend. This pronounced asymmetry suggests a consistent directional pattern whereby semantic divergence between ideological communities tends to increase over time, providing evidence for a linguistic manifestation of political polarization.

\begin{figure*}[h!]
\centering
\begin{minipage}[t]{0.255\textwidth}
\centering
\vspace{0pt}
\resizebox{1.0\columnwidth}{!}{
\begin{tabular}{lrc}
word & z-score & trend \\ \hline
secular    & 3.220  & $\uparrow$ \\
rifle      & 2.958  & $\uparrow$ \\
invasion   & 2.958  & $\uparrow$ \\
insurance  & 2.647  & $\uparrow$ \\
religion   & 2.647  & $\uparrow$ \\
victims    & 2.647  & $\uparrow$ \\
genocide   & 2.647  & $\uparrow$ \\
racists    & 2.491  & $\uparrow$ \\
economics  & 2.491  & $\uparrow$ \\
...        & ...    & ... \\
...        & ...    & ... \\
patriot    & 2.024  & $\uparrow$ \\ \hline
judge      & -2.491 & $\downarrow$ \\
hypocrisy  & -2.335 & $\downarrow$ \\
war        & -2.024 & $\downarrow$ \\
lawyers    & -2.024 & $\downarrow$ \\
boycott    & -1.968 & $\downarrow$ \\ \hline
campaign   & ...     & -- \\
fighting   & ...     & -- \\
voting     & ...     & -- \\
holocaust  & ...     & -- \\
propaganda & ...     & -- \\
abortion   & ...     & -- \\
\end{tabular}
}
\end{minipage}
\hfill
\begin{minipage}[t]{0.730\textwidth}
\centering
\vspace{0pt}
\includegraphics[width=\linewidth]{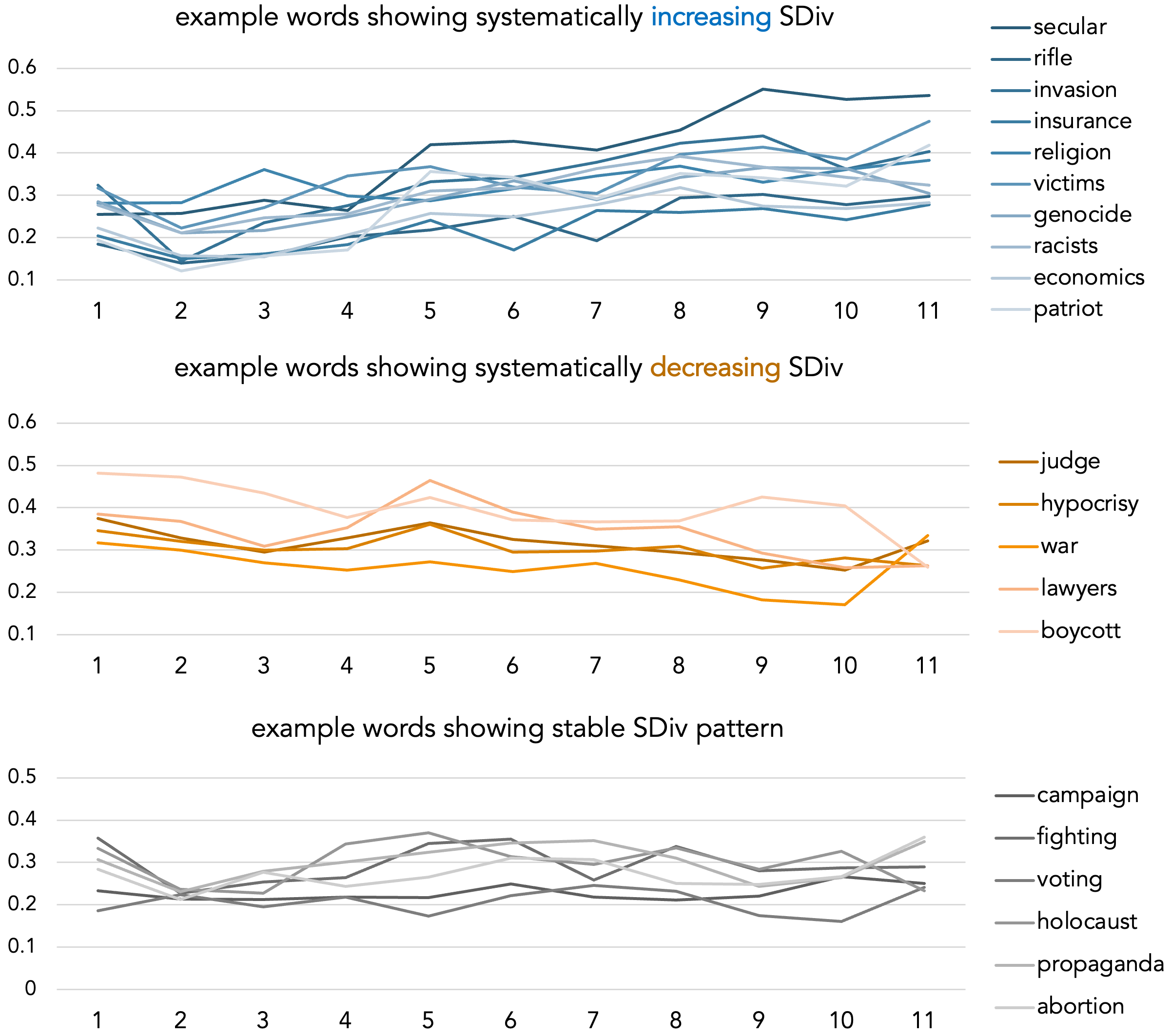}
\end{minipage}
\caption{Diachronic evolution of semantic diversity (\texttt{SDiv}) for example concepts over the 11 time periods in our study. The table lists words identified as exhibiting statistically significant upward~($\uparrow$) and downward~($\downarrow$) trends according to the Mann-Kendall test (p{<}0.05), as well as words showing no significant trend (--). \texttt{z-score} indicates the direction (and significance) of a monotonic trend. The plots illustrate corresponding temporal trajectories.}
\label{fig:temporal-trends}
\end{figure*}

Figure~\ref{fig:temporal-trends} (left) presents example words exhibiting increasing, decreasing, or no significant \texttt{SDiv} trends. Figure~\ref{fig:temporal-trends} (right) illustrates representative examples of increasing, decreasing, and stable \texttt{SDiv} trajectories across the 11-point time period.

\begin{figure}[h!]
\centering
\includegraphics[width=\columnwidth]{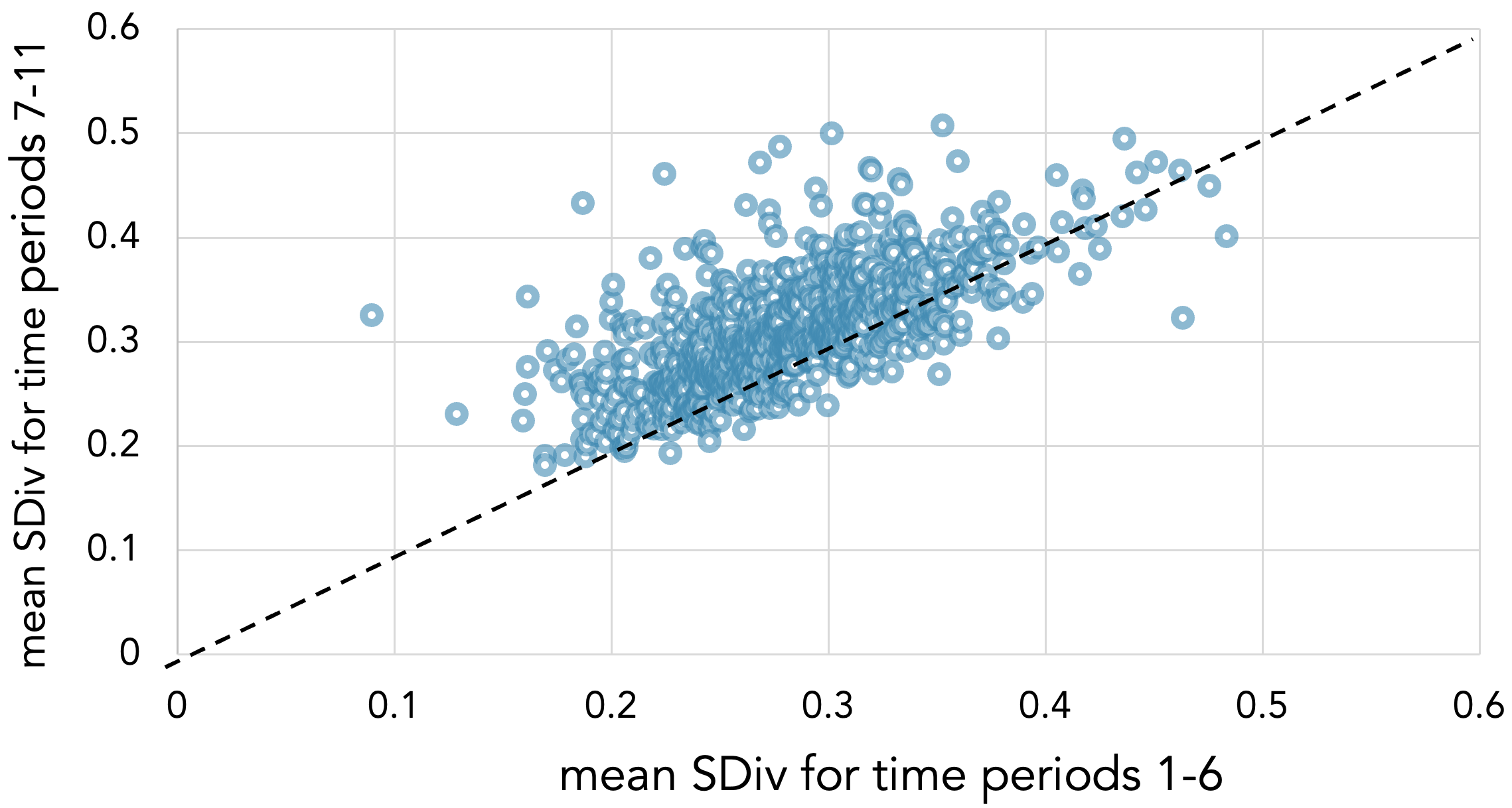}
\caption{Mean \texttt{SDiv} for the entire set of 850 words in our data for years 2008--2017 vs 2018--2023. The clear asymmetric pattern illustrates the increasing semantic divergence for the majority of concepts over time.}
\label{fig:sdiv-scatter}
\end{figure}

Additional evidence for increasing semantic divergence over time is presented in Figure~\ref{fig:sdiv-scatter}. The 11 time periods were divided into two halves: periods 1--6 (spanning 2008--2017) and periods 7--11 (spanning 2018--2023). For each word $w$, \texttt{mean(SDiv(w))} between the two communities was computed separately for the two halves. This provides a complementary, coarser-grained analysis of the overall temporal trend: the systematically higher average \texttt{SDiv} of a word during periods~{7--11} than during periods~{1--6} provides additional evidence for increasing divergence over time.

\vspace{-0.1cm}
\paragraph{Qualitative Analysis}
Increasing semantic divergence between the two communities calls for further interpretation. Careful inspection of a word's semantic neighborhood over time can shed light on the contextual shifts that cause its representations in the two communities to gradually diverge. Since manually inspecting each word across all 11 time periods is impractical, we simplify the analysis by using the two aggregated periods introduced in the previous section (2008--2017 and 2018--2023) and comparing each word's semantic neighborhood across these periods. Words exhibiting increasing \texttt{SDiv} are expected to display progressively diverging sets of semantic neighbors over time.

Table~\ref{tbl:cross-period-neigh} shows that the observed increase in semantic divergence is driven by gradual shifts in the contextual associations of words, reflected in increasingly distinct semantic neighborhoods across the two political communities.

\begin{table*}[h!]
\centering
\small

\begin{tabularx}{\textwidth}{l|c|X|X}
\multicolumn{2}{c}{} &
\multicolumn{1}{c|}{Republicans} &
\multicolumn{1}{c}{Democrats} \\

word &
\texttt{SDiv} &
\multicolumn{1}{c|}{top-K semantic neighbors} &
\multicolumn{1}{c}{top-K semantic neighbors} \\
\hline

secular (1) & 0.303 &
pluralistic, nonreligious, humanist, atheistic, humanists, pluralism, humanism, humanistic, secularized, prowestern, nondenominational, westernized, irreligious, theistic, theocratic &
humanist, doctrines, sects, judaism, orthodox, theocratic, theocracy, catholicism, nonreligious, denomination, judeochristian, zionism, fundamentalist, classically, individualist \\
\hline
\rowcolor{gray!10}
secular (2) & 0.377 &
\textcolor{red}{nonsecular}, humanists, \textcolor{red}{theistic}, \textcolor{red}{secularized}, atheistic, \textcolor{red}{irreligious}, \textcolor{red}{nontheistic}, nonreligious, \textcolor{red}{theism}, humanism, humanist, \textcolor{red}{religious}, humanistic, \textcolor{red}{christian}, \textcolor{red}{episcopalians} &
anglosaxon, \textcolor{blue}{denominations}, \textcolor{blue}{faiths}, philosophers, egalitarian, \textcolor{blue}{protestants}, \textcolor{blue}{orthodox}, \textcolor{blue}{catholicism}, \textcolor{blue}{judeochristian}, \textcolor{blue}{protestant}, espoused, \textcolor{blue}{theology}, \textcolor{blue}{nonchristian}, homogeneous, \textcolor{blue}{westboro} \\
\hline \hline

poverty (1) & 0.157 &
homelessness, illegitimacy, wedlock, singleparent, joblessness, destitution, squalor, outofwedlock, motherhood, fatherlessness, malnutrition, fatherless, fpl, selfperpetuating &
homelessness, subsistence, dependency, alleviate, incomes, stricken, generational, destitute, inequality, misery, obesity, starvation, abject, equilibrium, incidence \\
\hline
\rowcolor{gray!10}
poverty (2) & 0.171 &
destitution, impoverished, \textcolor{red}{singleparenthood}, homelessness, \textcolor{red}{singleparent}, inequality, starvation, \textcolor{red}{fatherlessness}, \textcolor{red}{wedlock}, squalor, \textcolor{red}{singlemother}, \textcolor{red}{singlemotherhood}, \textcolor{red}{fatherless}, poorer &
homelessness, stricken, \textcolor{blue}{criminalization}, calculator, incomes, \textcolor{blue}{stagnation}, \textcolor{blue}{lowerincome}, incidence, \textcolor{blue}{inequalities}, \textcolor{blue}{inequity}, alleviate, \textcolor{blue}{socioeconomic}, obesity, \textcolor{blue}{inequality}, underemployed \\

\end{tabularx}

\caption{Comparison of representative semantic neighbors in the first (\textbf{1}: 2008--2017) and the second (\textbf{2}: 2018--2023) period. Colored terms indicate second-half Republican and Democratic neighbors, illustrating the diachronic divergence. For "secular", the Republican neighborhood increasingly emphasizes the belief--unbelief distinction, whereas the Democratic neighborhood remains more closely associated with religions, denominations, and religious institutions. For "poverty", the Democratic neighborhood becomes more associated with inequality and socioeconomic conditions, while the Republicans remain more closely associated with family structure.}

\label{tbl:cross-period-neigh}
\end{table*}

\subsection{Topic-level Diachronic Polarization}
We next move from individual concept- to topic-level analysis, aiming to obtain more robust insights into ideological polarization between the two communities over time. We first group the 850 identified words into semantic clusters, and then perform divergence trend analysis over these groupings rather than individual words. Clusters exhibiting a steadily increasing trend in semantic divergence are indicative of a widening gap in the way topics are discussed across the communities.

\begin{table*}[h!]
\centering
\resizebox{\textwidth}{!}{
\begin{tabular}{l|c|c|p{10cm}}
cluster name & size & trend & sample words in a cluster \\ \hline
political ideologies and beliefs & 20 & $\uparrow$ & socialist, leftist, capitalism, conservative, liberal, ...\\
legal and political status & 18 & $\uparrow$ & supreme, legal, illegally, constitutional, judicial, racially, elite, ...\\
violence and defense actions & 17 & $\uparrow$ & attack, threat, invasion, damage, assaulted, violate, defend, ...\\
crime and law enforcement & 17 & $\uparrow$ & gang, police, mob, riot, riots, rioting, cops, looting, ...\\
social and legal status & 14 & $\uparrow$ & eligible, controlled, oppressed, allowed, mandated, woke, ...\\
guns and weapons violence & 12 & $\uparrow$ & guns, weapons, shooter, drugs, shooting, rifle, firearms, ...\\
LGBTQ+ and gender issues & 10 & $\uparrow$ & homosexuality pedophilia, gay, trans, transgender, gender, ...\\ 
crime and justice & 10 & $\uparrow$ & criminals, innocent, crime, victim, trials, felon, ...\\ \hline
insults and negative criticism & 19 & -- & vile, incompetence, ignorance, idiot, garbage, corrupt, dumb...\\ 
titles and roles in government & 15 & -- & senator, citizen, attorney, chief, judge, mayor, officer, president, ...\\ 
protests and public outrage & 13 & -- & protests, arrest, criticism, boycott, outrage, allegations, ...\\ 
\end{tabular}
}
\caption{Semantic word clusters exhibiting significantly growing and stable mean \texttt{SDiv} diachronic pattern: all 8 clusters with increasing \texttt{SDiv} at p{<}0.05 ($\uparrow$), and 3 example clusters with stable trend (--). Increasing semantic divergence is concentrated in substantive policy and identity-related topics, whereas more institutional or generic discourse categories remain largely stable over time.}
\label{tbl:cluster-sdiv-results}
\end{table*}

\subsubsection{Extracting Topical Clusters}
All 850 politically associated words used in this study naturally form several thematic clusters of varying sizes. For example, the words \textit{voting}, \textit{electorate}, \textit{election}, \textit{campaign}, \textit{candidate}, \textit{agenda}, and \textit{primaries} make up a cluster referring to \textit{elections and voting}. Clusters may vary considerably in size, and their number is unknown in advance, and therefore has to be discovered automatically. Moreover, some words may not have a sufficient number of similar counterparts to form a cluster and should therefore be treated as outliers. We use the clustering algorithm proposed by \citet{rabinovich2022gaining}, which is specifically tailored to this setting, to group words by their meaning, using the \texttt{intfloat/e5-large-v2} encoder,\footnote{\url{https://huggingface.co/intfloat/e5-large-v2}} a similarity threshold of 0.810,\footnote{The similarity threshold was tuned through manual inspection of the produced clusters; varying it slightly (within the [0.805, 0.815] range) changed the number of clusters (and their density), while not affecting the final findings.} and a minimum cluster size of 10. This setting resulted in 37 clusters covering 563 words (out of 850). We further used the GPT-5.1-mini model \citep{openai_gpt51_2026} to assign meaningful names to the generated clusters.

\subsubsection{Diachronic Semantic Divergence}
A cluster's diachronic semantic divergence was computed as the mean period-wise \texttt{SDiv} over its individual words in two ways: (1) assigning uniform weights to all words within a cluster, and (2) weighting each word proportionally to its frequency in the dataset. Both approaches yielded similar results; we therefore adhere to the simpler (uniformly weighted) method hereafter.

Out of the 37 produced clusters, 8 were detected as exhibiting a significantly increasing \texttt{SDiv} trend using the Mann--Kendall test \citep{mann1945nonparametric} at a significance level of p{<}0.05. No topical clusters exhibited a significant trend in the opposite direction; that is, no grouping of words became semantically closer over time across the two communities. Notably, of the 37 fitted trend lines, 36 had a positive slope and only one had a negative slope: 31 out of 37 groupings exhibit increasing semantic divergence (albeit not statistically significant) between the left- and right-wing communities over time.
Example groupings and their temporal trends are presented in Table~\ref{tbl:cluster-sdiv-results}: increasing semantic divergence is concentrated in substantive policy and identity-related topics, while more institutional or generic discourse categories remain largely stable.

\section{Conclusions}
\label{sec:conclusions}

In this study we presented a longitudinal analysis of ideological polarization in Reddit political discourse using semantic divergence between community-specific word embeddings. Our results show that ideological polarization has increased over the past fifteen years, both for individual political concepts and broader topical clusters, reflecting increasingly distinct ideological framing. More broadly, the proposed framework provides a scalable approach for studying the temporal dynamics of semantic polarization in various forms of online discourse, and the released resources would facilitate further research in this field.

\section{Limitations}
\label{sec:limitations}

One limitation of this study concerns the relatively small number of temporal observations available for trend analysis. Although the 15-year corpus spans a substantial period of time, the data had to be aggregated into only 11 time periods in order to ensure sufficiently large corpora for reliable embedding training. Such a small number of observations is not ideal for statistical trend detection, reducing the power of the Mann--Kendall test and increasing the likelihood of false negatives. Consequently, while we identified a substantial number of words exhibiting statistically significant semantic divergence over time, additional significant trends would likely emerge if larger corpora permitted a finer-grained temporal partitioning.

A second limitation stems from the nature of the underlying data. Our analysis is based exclusively on discussions from politically oriented Reddit communities, whose participants may be more politically engaged and ideologically committed than the general population. As a result, the observed patterns should not necessarily be interpreted as representative of political discourse in society at large. Nevertheless, Reddit provides a unique source of large-scale, longitudinal, and authentic user-generated political discourse, making it well suited for studying the evolution of ideological language over time.

\section*{Ethical Considerations}
\label{sec:ethical}

Here we address the main concern of anonymity of Reddit users. Data used for this research can only be associated with participants' user IDs, which, in turn, cannot be linked to any identifiable information, or used to infer any personal or demographic trait. \citet{jagfeld2021understanding} debated the need to obtain informed consent for using social media data mainly because it is not straightforward to determine if posts pertain to a public or private context. Ethical guidelines for social media research \citep{benton2017ethical} and practice in comparable research projects \citep{ahmed2017using}, as well as {Reddit's terms of use},\footnote{\url{https://www.redditinc.com/policies/user-agreement-september-12-2021}} regard it as acceptable to waive explicit consent if users' anonymity is protected.
We reinforce that our dataset does not contain user IDs for neither posts nor comments. This data can be retrieved using a post or comments ID, which is attached to each text in the dataset.


\bibliography{custom,anthology-1,anthology-2}



\end{document}